\documentclass[10pt, a4paper]{article}
\usepackage{lrec-coling2024} 
\usepackage{multirow} 
\usepackage{array}
\usepackage{booktabs}
\usepackage{amsmath}
\usepackage{mathptmx}
\usepackage{mathtools}

\title{ \textbf{PRIMO: Progressive Induction for Multi-hop Open Rule Generation}}

\name{Jianyu Liu, Sheng Bi$^{*}$\thanks{$^{*}$ Corresponding author}, Guilin Qi} 

\address{Southeast University, Nanjing, 211189, Jiangsu, China\\
\{liujianyu, bisheng, gqi\}@seu.edu.cn}

\abstract{
Open rule refer to the implication from premise atoms to hypothesis atoms, which captures various relations between instances in the real world. Injecting open rule knowledge into the machine helps to improve the performance of downstream tasks such as dialogue and relation extraction. Existing approaches focus on single-hop open rule generation, ignoring multi-hop scenarios, leading to logical inconsistencies between premise and hypothesis atoms, as well as semantic duplication of generated rule atoms.
To address these issues, we propose a progressive multi-stage open rule generation method called PRIMO. We introduce ontology information during the rule generation stage to reduce ambiguity and improve rule accuracy. PRIMO constructs a multi-stage structure consisting of generation, extraction, and ranking modules to fully leverage the latent knowledge within the language model across multiple dimensions. Furthermore, we employ reinforcement learning from human feedback to further optimize model, enhancing the model's understanding of commonsense knowledge. Experiments show that compared to baseline models, PRIMO significantly improves rule quality and diversity while reducing the repetition rate of rule atoms.
 \\ \newline \Keywords{Open rule, Pre-trained language model, Reinforcement learning from human feedback} }

\begin{document}

\maketitleabstract

\section{Introduction}

Rules usually refer to objective regularities or logical relationships of domain concepts, usually expressed in the form of ``if-then'' statement~\cite{DBLP:journals/fss/NovakL06}. Rules can describe most complex knowledge, while naturally incorporating domain-specific knowledge~\cite{DBLP:journals/eswa/Chi10}. When reasoning based on rule, users can intuitively understand the process logically \cite{wason1968reasoning}. Therefore, rules are widely used in downstream applications, such as intelligent data analysis \cite{becquet2002strong} and knowledge discovery \cite{DBLP:journals/widm/Garcia-VicoCMGJ18}. For example, Lin et al. \shortcite{DBLP:journals/nle/LinP01} suggested that rule-based reasoning can quickly narrow the search space in question answering.

Rule generation aims to discover rules that satisfy logical constraints from large amounts of data. Traditional research has been devoted to generating rules by observing data commonalities. For example, one of the core tasks of Inductive Logic Programming (ILP)~\cite{DBLP:journals/ai/Muggleton99} is to mine rules in the form of Horn clauses from data \cite{DBLP:conf/alt/RaedtK04}. Since the axioms of the rules are restricted to the entities and relations already present in the given context, this leads to a limited and fragile expression of such rules \cite{wrobel2001inductive}. Furthermore, these methods have weak generalization capabilities due to the scale of the knowledge source \cite{DBLP:journals/ai/Muggleton99}. 

In recent years, some researchers proposed open rule generation, aiming to generalize more diverse rules from large-scale open KBs. Hwang et al. proposed COMET \shortcite{DBLP:conf/aaai/HwangBBDSBC21}, a pre-trained language model that can learn commonsense from natural language. Given any text, COMET can generate new rules of the form of If-Then statement. However, COMET's training dataset, ATOMIC2020 \cite{DBLP:conf/aaai/SapBABLRRSC19}, contains only 23 manually defined relations, which limits the types of rules that can be generated. Orion \cite{DBLP:conf/nips/CuiC21} adopts an unsupervised approach to utilize the knowledge in the Pre-trained Language Models (PLM) to automatically mine open rules. However, since it does not consider the ontological information of the entities in the rules, Orion tends to generate rules that are not logically self-consistent.
Existing approaches focus on single-hop open rule generation, i.e. the generation of multiple parallel hypothesis atoms based on given premise atoms. However, it is difficult to extend to some complex scenarios due to the short chain of rule reasoning and weak expression of complex logic capabilities, such as multi-round dialogues. In multi-hop open rule generation, the currently generated hypothesis atom must take into account all previously generated rule atoms, which places higher demands on the model's reasoning ability. In addition, multi-hop open rule generation requires the model to own global information awareness. Existing methods do not have long-term context awareness, which leads to logical incoherence between atoms.

To address these issues, we propose PRIMO --- a \underline{P}rog\underline{R}essive multi-stage \underline{I}nduction method for \underline{M}ulti-hop \underline{O}pen rule generation. By introducing ontological information of entities into the hypothetical atom generation procedure, the generation of incorrect rules can be effectively mitigated. Considering the reasoning challenges of multi-hop open rules, we introduce generation, extraction and ranking modules in each sub-rule generation phase. Multiple modules are connected through the designed prompt, and the modules collaborate with each other to progressively generate multi-hop open rules. To reduce the repeated generation of rule atoms, we update the prompt after each derivation to learn the prior information of the generated atoms. Finally, after fine-tuning the model, we construct reward signals based on human feedback, which further enhance reasoning with common sense through reinforcement learning.

To evaluate the effectiveness of multi-hop open rule generation, we constructed a benchmark dataset and evaluated various systems using a wide range of automated metrics and human judgement.The results show that PRIMO effectively improves performance by splitting rule generation into multiple stages. Moreover, thanks to the stage-wise updating strategy of prompts, our approach significantly reduces the generation of repetitive atoms. It outperforms a series of baseline models and achieves performance close to LLM, confirming the effectiveness and superiority of PRIMO.

\section{Related Work}
\textbf{Text generation based on PLM} \quad  
Early research on text generation primarily relied on manual rules and predefined templates~\cite{DBLP:conf/emnlp/KaleR20}. These methods involved creating templates for text generation by manually extracting features and applying some simple syntactic and grammatical rules to organize the generated text. Rule-based and template-based approaches required a significant amount of manual effort and were limited in their application due to fixed generation patterns and narrow use cases.
With the development of deep learning, models such as Recurrent Neural Networks (RNN) \cite{DBLP:conf/emnlp/ChoMGBBSB14}, Long Short-Term Memory Neural Networks (LSTM) \cite{DBLP:journals/tnn/GreffSKSS17}, and Transformer~\cite{DBLP:conf/nips/VaswaniSPUJGKP17} have brought significant performance improvements in text generation.
Recently, text generation methods based on pre-trained language models have demonstrated unparalleled performance, capable of generating fluent text in few-shot and zero-shot scenarios~\cite{iqbal2022survey}.

\noindent \textbf{Open rule generation} \quad Open rule generation involves summarizing and inducing rules from an open knowledge base (KB), which means deriving new information from known information. Previous work has focused on discovering rules within systems, and traditional rule generation methods typically rely on closed datasets, such as Inductive Logic Programming (ILP) \cite{DBLP:conf/alt/RaedtK04}, AMIE\cite{DBLP:conf/www/GalarragaTHS13}, AMIE+\cite{DBLP:journals/vldb/GalarragaTHS15}, and other methods. These methods have limited expressive power since they lack common sense, and their rules are restricted to existing entities and relations. John McCarthy \shortcite{McCarthy1984SomeES} suggested that these rules lack commonsense and are ``difficult to extend beyond the scope originally contemplated by their designers''. In recent years, researchers have discovered that PLM can serve as high-quality open KBs \cite{DBLP:conf/emnlp/PetroniRRLBWM19,DBLP:journals/corr/abs-2010-11967}and commonsense KBs\cite{DBLP:conf/aaai/SapBABLRRSC19,DBLP:journals/corr/abs-1806-02847}. COMET \cite{DBLP:conf/aaai/HwangBBDSBC21} was trained on annotated If-Then rule sets, extending the knowledge representation from structured KBs to open natural language knowledge. 
However, COMET is limited by the type of rules for a given KB, which deviates from the principle of summarizing data commonalities to generate rules, resulting in a weaker expression.
Cui et al.~\shortcite{DBLP:conf/nips/CuiC21} proposed an unsupervised rule generation method called Orion, indicating that PLM can be leveraged as a KB to discover commonalities in data. However, Orion ignores the ontological information associated with the rules. As a result, it may generate rules that are logically inconsistent.

\section{Problem Definition}
Rules employ logical symbols to describe implicit concepts or patterns in data. This paper references the definition of an open rule based on Horn clauses~\cite{DBLP:conf/emnlp/SchoenmackersDEW10}. In Horn clauses, atoms are facts that may contain variables in the subject and/or object~\cite{DBLP:conf/emnlp/SchoenmackersDEW10}. Cui et al. \shortcite{DBLP:conf/nips/CuiC21} define an open rule as an implication from a premise atom to a hypothesis atom. We will follow the same definition in this paper. The formal definition of an open rule is given in the following form:

\noindent \textbf{Definition 1} (Open rule). The open rule is a logical deduction from the premise atom $(x,\ r_p,\ y)$ to the hypothesis atom $(x,r_h,\ y)$:
\begin{equation}
    (x,{{r}_{p}},y) \to (x,{{r}_{h}},y)
\end{equation}
where $r_p$ and $r_h$ represent relations described in natural language. This rule suggests that if entity $x$ and $y$ have $r_p$, they are also likely to have $r_h$. In contrast to rules with strict formal definitions, open rules have a more expressive form, which makes them better suited to capture the complexities of the real world.
For example, the open rule $(x, was\ born\ in, y) \rightarrow (x, is\ citizen\ of, y)$ means that a person being born in a particular country (usually) is a citizen of that country. And $(Kobe, was\ born\ in, USA) \rightarrow (Kobe, is\ citizen\ of, USA)$ is an instance of the rule mentioned above.

Given a premise atom $(x, r_p, y)$, we aim to induce one most relevant hypothesis atom from LM. Specifically, we define the problem as follows:

\noindent \textbf{Definition 2} (Open rule generation). For a given premise atom $(x, r_p, y)$, find the $r_h$ with the highest probability $P(r_h \mid r_p)$ ,and then obtain the hypothesis atom $\left(x, r_h, y\right)$.
In an open rule, the number of logical deductions from the premise atom is called its hop number. Single-hop open rules have a weaker logical reasoning capability due to shorter reasoning chains. In practical application scenarios, there is a greater need for multi-hop open rules.

\noindent \textbf{Definition 3} (Multi-hop open rule generation). Starting from the premise atom, deduce forward in sequence to generate a series of hypothesis atoms that form a chain of rules:
\begin{equation}
    (x,{{r}_{p}},y)\to (x,{{r}_{h}}_{1},y)\to (x,{{r}_{h}}_{2},y)\to \dots
\end{equation}

The challenges of multi-hop open rule generation are mainly in two aspects:

(1) Logical Inconsistency: Unlike single-hop rule generation, where only one logical deduction is performed, in multi-hop rule generation, each hop generation must consider all previously generated rule atoms. The complexity of inference will increase due to the above requirement. Ensuring logical coherence is critical to the task, and contradictions among rule atoms should be avoided.

\begin{figure*}[t]
    \centering
    \includegraphics[width=0.85\linewidth]{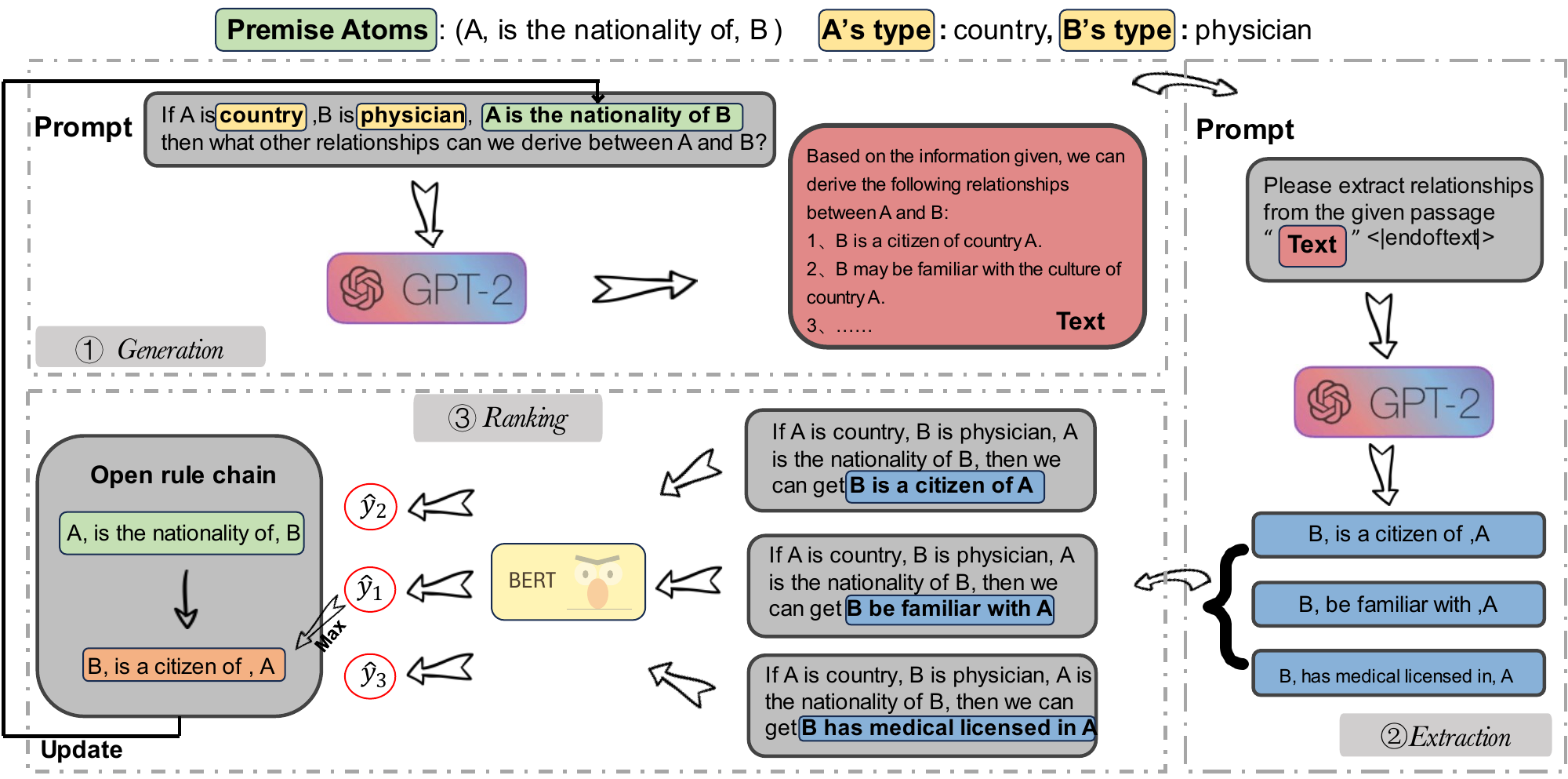}
    \caption{PRIMO consists of three modules. \textit{Generation} module creates descriptions of hypothesis atoms based on the premise atoms. \textit{Extraction} module extract atoms implied in text output of the generation stage. \textit{Rank} module evaluates the plausibility of candidate hypothesis atoms.} %
    \label{figure1}
\end{figure*}

(2) Semantic Repetition: During multi-hop generation, generating atoms with semantic repetition compared to previously generated atoms is undesirable, as such they do not provide any valuable information. Therefore, the semantics of each atom should be diversified to reduce semantic repetition.

\section{PRIMO}
As shown in Figure~\ref{figure1}, PRIMO consists of three stages, i.e., \textit{\textbf{Generation}, \textbf{Extraction}}, and \textit{\textbf{Ranking}}, connected sequentially. The \textit{Generation} module creates descriptions of hypothesis atoms based on the premise atoms. The \textit{Extraction} module retrieves hypothesis atoms implied in the text output of the generation stage, and the \textit{Ranking} module evaluates the plausibility of candidate hypothesis atoms. The design reasons for the progressive framework can be summarized as follows:

(1) We observed that adopting an end-to-end approach, where the model directly generates hypothesis atoms, makes it challenging to explore a logically coherent rule chain and tends to generate repetitive rule atoms. By combining three small-scale language models, each model serving a different purpose, and refining the reasoning process, we can achieve better generation performance. By dividing the process into distinct stages – Generation, Extraction, and Ranking – we believe that a multi-stage approach is more controllable and transparency in rule generation compared to a single-stage method, as we can directly observe the output of each stage and adjust the model for a particular stage individually.

(2) Each module is independent and exhibits good transferability. PRIMO utilizes different base models in each stage, enabling these models to complement each other’s strengths and weaknesses. At the same time, the flexibility of the framework allows the underlying model to be changed at any stage as needed to improve overall performance, rather than relying entirely on the performance of any singlone model. By fine-tuning the models separately for different domains of data, we enhance the overall network's transferability.
We will discuss thprovide details of the architecture from Sec.~\ref{generation} to Sec.~\ref{multi-hop-generation}.

\subsection{Generation}
\label{generation}
Due to the substantial implicit common sense contained within PLM, this work aims to establish a high-performance open rule generation method that can be achieved without manual annotation.

Cui et al~\shortcite{DBLP:conf/nips/CuiC21} suggests that the lack of ontology information about entities may lead to semantic conflicts between premise and hypothesis atoms. For instance, given \texttt{[X] is a provincial capital of [Y]}, Orion may output some atoms like \texttt{[X] is a river of [Y]}. Therefore, we attempt to introduce entity type information into the open rule generation process to improve the correctness and diversity of rule. Because adding type information can provide ontology-level constraints, we provide it to the \textit{Generation} module for a given atomic entity pair.

Initially, we input a prompt into the PLM. As shown in Figure~\ref{figure1}, the prompt for the \textit{Generation} module is constructed as follows: ``\texttt{If A is $\operatorname{typeA}$, B is $\operatorname{typeB}$, $\operatorname{premise \ atoms}$, then what other relationships can we derive between A and B?}'' It contains three slots, $\operatorname{typeA}$ and $\operatorname{typeB}$, which are filled with the type information of entity A and entity B. And $\operatorname{premise \ atoms}$, which consists of the given premise atom. Importantly, during the subsequent multi-hop generation, in addition to the provided premise atom, $\operatorname{premise \ atoms}$ need to be updated to include the atoms generated in the previous generations.

The goal of the \textit{Generation} module is to induce the model to ``speak out'' its internal implicit knowledge commonalities based on the initial information provided by the prompt, thereby generating rule atoms. Therefore, through a ``dialogue'' with the prompt, the \textit{Generation} module is tasked with describing other potential relations that may exist between entity pairs. Taking into account the nature of the task, we choose GPT-2, which is well suited forexcels at generative tasks, as the base model for the \textit{Generation} module.

\subsection{Extraction}
As previously stated, the \textit{Generation} module performs reasoning on potential relations between pairs of entities based on the provided premise atoms. The output text from the \textit{Generation} module contains a wealth of reasoning knowledge, including an analysis of entity pair type information, descriptions of the establishment of premise atoms, and scenarios of other possible relations that may exist between entity pairs. We need to summarize and extract the key information from the output of the \textit{Generation} module. Therefore, we design the \textit{Extraction} module to accomplish this task.

As shown in Figure~\ref{figure1}, the \textit{Extraction} module first fills the \textbf{Text} generated by the \textit{Generation} module into predefined slots in a prompt and then feeds this prompt to the model. After fine-tuning, the model learns to extract hypothesis atoms from the given text. Therefore, the model outputs a set of candidate hypothesis atoms.

Similar to the \textit{Generation} module, we choose GPT-2 as the base model for the \textit{Extraction} module. Although the prompt aims to instruct GPT-2 to extract only hypothesis atoms from the given text, we found that the model may still output some atoms that do not exist in the given text. We attribute this to GPT-2's limitation in ensuring factual accuracy. To achieve better extraction performance, we further optimize GPT-2 using reinforcement learning from human feedback, as described in Sec.~\ref{RLHF}.

\subsection{Ranking}
As described above, we have obtained a set of candidate hypothesis atoms in the first two stages. However, these atoms may have problems such as logical inconsistency and semantic repetition with existing atoms. In the next stage, we need to evaluate the plausibility of these generated atoms.

First, as shown in Figure~\ref{figure1}, each candidate hypothesis atom is filled into a statement that is to be evaluated. This statement follows the format: ``\texttt{If A is $\operatorname{typeA}$, B is $\operatorname{typeB}$, $\operatorname{premise \ atoms}$, we can get $\operatorname{hypothesis \ atom}$}.'' A total of $n$ statements are generated, where $n$ is the number of hypothesis atoms output by the \textit{Extraction} module. Next, these statements are fed into a well-trained Bert. Bert encodes and scores these $n$ statements, and the hypothesis atom with the highest score is considered the most reliable and is added to the open rule chain. The goal of Bert is to map an input text sequence to a reward value, which numerically corresponds to human preferences.

For Bert's training, we use ranked sequences of open-rule statements, rather than artificially scoring the statements directly, in order to mitigate the potential noise caused by variations in annotators' perspectives and to reduce the bias introduced by subjective human scoring. For example, if there are four hypothesis atoms and their order is $A > B > C > D$, and we need to train a scoring model so that it can assign appropriate scores to the four atoms based on the order, i.e., $r(A) > r(B) > r(C) > r(D)$. Here $r()$ represents the score of the atom. Therefore, the loss function is designed as follows:
\begin{equation}
    loss(\theta )=-\sum\limits_{w<\ell}{\left[ \log \left( \sigma \left( {{r}_{\theta }}\left( {{y}_{w}} \right)-{{r}_{\theta }}\left( {{y}_{\ell}} \right) \right) \right) \right]}
\end{equation}
where ${y_w}$ represents all the atoms ranked above ${y_\ell}$. To better normalize the differences, we applied a sigmoid function to each pairwise difference to bring the values into the range of 0-1. After training, the model's scoring of hypothesis atoms reflects human value judgements, allowing the plausibility of atoms to be assessed.

\subsection{Multi-hop Open Rule Generation}
\label{multi-hop-generation}
The multiple-hop generation of the open rule can be viewed as consisting of $n$ successive single-hop generations. Through the \textit{Generation-Extraction-Ranking} process, we can obtain the hypothesis atom that the model deems most reasonable given the current information. Then, as shown in Figure~\ref{figure1}, we record the top-ranked generated atom and add it to the end of the open rule chain. For the next hop of rule generation, the hypothesis atom generated in the previous hop should serve as the premise atom for the current hop's rule generation. So the \textit{Generation} module needs to consider all the premise atoms in the chain. We update the \textit{Generation} module's prompt with the premise atoms along the chain, starting with the initial atom and concatenating all the atoms in the order they were generated. The prompt of the \textit{Extraction} module remains unchanged and continues to output the candidate hypothesis atoms extracted from the text generated by the \textit{Generation} module. The statements inputted into \textit{Ranking} must also update the contents of the premise atoms. Hence, the open rule chain is continuously updated, repeating the single-hop generation incrementally to generate multi-hop open rules until the chain reaches the predefined length. Since both the \textit{Generation} and \textit{Ranking} modules have global information updates, PRIMO combines the information from all known atoms when completing the next hop generation. This helps ensure logical consistency between atoms and reduces semantic repetition.

\subsection{RLHF}
\label{RLHF}
ChatGPT proposed by OpenAI~\cite{openai} breaks the boundaries between machines and humans. This innovative model excels in various domains of tasks. The underlying work is based on a novel training paradigm in the field of Large Language Models (LLMs), namely Reinforcement Learning from Human Feedback (RLHF)~\cite{DBLP:conf/nips/Ouyang0JAWMZASR22}. Over the past few years, the ability of various LLMs to generate diverse text based on prompts has been quite impressive. However, the evaluation of the generated results is subjective and context dependent, making it challenging to measure these results using existing text generation metrics such as BLEU~\cite{DBLP:conf/acl/PapineniRWZ02} and ROUGE~\cite{Lin2004ROUGEAP}.

For instance, the goal of open rule generation is to produce genuine information. Existing methods typically rely on next-word prediction and simple loss functions, such as cross-entropy. However, these methods lack the explicit incorporation of human preferences and subjective opinions. RLHF, on the other hand, employs reinforcement learning to optimize the model directly based on human feedback, encouraging the model's output to align with complex human values. To further enhance the rationality of the rule generated by PRIMO and to improve the performance of each stage in the model, we use RLHF for additional optimization.

For the \textit{Generation} module, our objective is to ensure that the generated text adheres to factual correctness as much as possible while avoiding redundancy with known information. For the \textit{Extraction} module, our goal is to ensure that the extracted results are faithful to the original text, matching with the descriptions in the text generated by the \textit{Generation} module, without introducing unrelated information. Since the \textit{Ranking} module is pre-trained with annotated data, its scoring results are to some extent a reflection of human value preferences. Therefore, it serves as the reward model for the \textit{Generation} module in RLHF. Considering different optimization objectives, for the \textit{Extraction} module, we employ human judgment to directly score the results in RLHF.

We utilize Proximal Policy Optimization (PPO) \cite{DBLP:journals/corr/SchulmanWDRK17} to optimize the initial PLM parameters. During the training of the \textit{Generation} module, the input and output are aligned with the configuration in Figure~\ref{figure1}. Subsequently, the \textit{Extraction} module performs extraction and the reward value of the \textit{Generation} module is the maximum score among all the hypothesis atoms evaluated by the \textit{Ranking} module. In the training of the \textit{Extraction} module, we fill the text generated by the \textit{Generation} module into the prompt, and then judge the output result by human scoring. Based on the score, we use PPO to optimize the model and complete the training for one data point. The updated the \textit{Extraction} module then proceeds to the training of the next data point, continuously optimizing the model. To prevent the optimization from getting out of control, we introduce $\mathrm{KL}$ divergence \cite{DBLP:journals/tit/ErvenH14} as a constraint on the objective function for the optimization:
\begin{equation}
    r={{r}_{\theta }}-\lambda {r}_{KL}
\end{equation}
where ${r_{\theta}}$ is the reward, and ${r_{KL}}$ calculates the $\mathrm{KL}$ divergence between the model's outputs before and after the update, which serves as a penalty term. In this way, we enable PRIMO to better adapt to human preferences, resulting in rules that more closely resemble common-sense cognition.

\section{Experiment}
\subsection{Dataset}
In order to evaluate the effectiveness of multi-hop open rule generation, we construct our benchmark dataset which contains 495 premise atoms. To build premise atoms that describe $x$ and $y$, we collect relations for pairs of entities from Freebase~\cite{DBLP:conf/sigmod/BollackerEPST08}. First, we need to use SPARQL queries~\cite{DBLP:conf/wise/HarrisS05} on Freebase~\cite{DBLP:conf/sigmod/BollackerEPST08} to find entities with the properties name (type.object.name) and type (common.topic.notable.types). However, we found that the entities retrieved in this way may still contain some conceptual entities rather than physical entities. To address this, we add the constraint properties (common.topic.description) to the query statement. Then, starting from the retrieved entities, we query entities that also have the three properties, and are connected to the first entity by the relation $entity1 \rightarrow entity2$. The former is denoted as $entity1$, the latter as $entity2$. We record their respective name, type, and the relation $entity1 \rightarrow entity2$.

\begin{figure}[t]
    \centering
    \includegraphics[width=0.8\columnwidth]{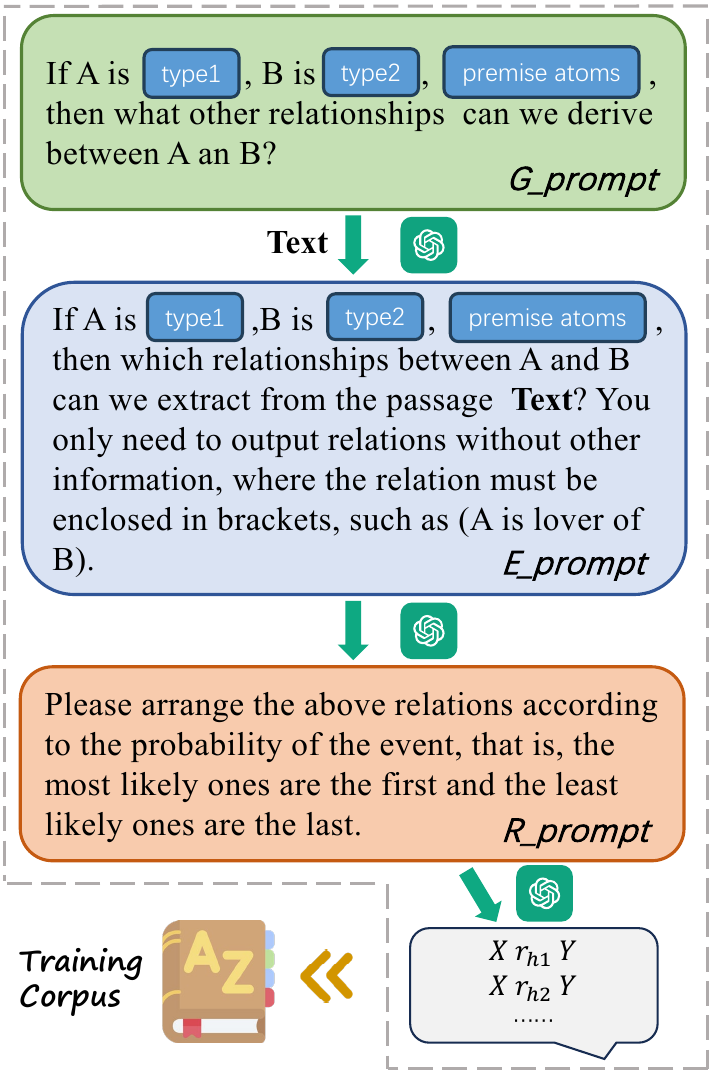}
    \caption{We construct dataset and collect training corpus by ChatGPT with three step. Step1: use G\_prompt to instruct ChatGPT to generate text that describes the relation between two entities. Step2: text from Step 1 is filled into E\_prompt to extract hypothesis atoms. Step3: Ranking these hypothesis atoms through R\_prompt.}
    \label{figure2}
\end{figure}

ChatGPT has sparked extensive discussions in the NLP community and society at large. It can engage in natural and fluent conversations with users, providing intelligent question-answering and text generation capabilities for various scenarios. Therefore, in this work, we want to leverage ChatGPT's powerful text generation capabilities  to achieve an AI self-sufficiency training mode. As shown in Figure~\ref{figure2}, we start by querying ChatGPT through prompt to obtain Text. Each of the three placeholders is filled with the above mentioned information extracted from Freebase~\cite{DBLP:conf/sigmod/BollackerEPST08}. Note that the premise atoms placeholder can hold multiple atoms for multi-hop open rule generation. In the second round of questioning, we fill the prompt with the text generated by ChatGPT in the first round. Then, in the third round of questioning, we ask ChatGPT to rank the candidate entity relations mentioned in the answer of the second round, based on the probability of their of their actual occurrence. The prompt for the three rounds of questioning is as shown in Figure~\ref{figure2}.

The text generated from these three rounds of questioning serves as the training corpus for the three modules of PRIMO. After the three rounds of questioning, the highest ranked atom in the last round is taken as a hypothesis atom. Then, we repeat the above process until the number of collected hypothesis atoms reaches the predetermined hop. Ultimately, to create an open rule chain for that data point, these atoms are sorted in the order in which they were generated.

In the end, we obtain an open rule multi-hop generation dataset consisting of 495 premise atoms, totaling 2851 samples. Each sample consists of three parts: premise atoms, entity types, and the open rule chain. The number of hops of the open rule chain ranges from 1 to 5.

\subsection{Baselines}
We use the following LM-based baseline models to evaluate the performance on the multi-hop open rule generation:


(1) \textbf{Prompt} We use a prompt ``$if\ r_p\ then\ [\mathrm{MASK}]$'' and make predictions using fine-tuned T5~\cite{DBLP:journals/jmlr/RaffelSRLNMZLL20}. The generated results serve as the rule hypothesis atom and are added to the premise atom information. Repeat the above process until the desired rule's hop is reached.

(2) \textbf{COMET}~\cite{DBLP:conf/aaai/HwangBBDSBC21} takes a premise atom as input and produces a collection of single-hop hypothesis atoms with different relations. 

(3) \textbf{Orion}~\cite{DBLP:conf/nips/CuiC21} summarizes the commonalities of a set of instances based on a given premise atom, and these commonalities are used to induce new open rules.

(4) \textbf{Vicuna}(13B)~\cite{chiang2023vicuna} is a large language model that is fine-tuned based on LLaMA~\cite{DBLP:journals/corr/abs-2302-13971}. 

\begin{table}[htp]
  \centering
  
   \resizebox{\columnwidth}{!} %
  { 
  \renewcommand{\arraystretch}{1.2} %
  \setlength\tabcolsep{2.5pt} %
  \begin{tabular}{cccccc}
        \hline
        Our Dataset & B1 & B2 & B4 & RL & Self-B2\\
        \hline
        Prompt & 33.0 & 13.2 & 0.1 & 39.2 & 89.2\\
        COMET & 35.1 & 13.6 & 1.1 & 42.9 & 92.6\\
        Orion & 39.9 & \textbf{19.5} & 0.1 & 52.5 & 86.4\\
        \hline
        Vicuna-13B & \textbf{44.8} & 17.4 & \textbf{2.9} & \textbf{67.9} &
        75.5\\
        \hline
        PRIMO & 44.3 & 16.5 & 2.1 & 66.3 & 80.5\\
        PRIMO-without RLHF & 42.5 & 15.0 & 2.0 & 64.5 & 77.7\\
        PRIMO-train G\_Net & 43.4 & 14.1 & 1.3 & 66.3 & 77.8\\
        PRIMO-train E\_Net & 40.7 & 15.1 & 2.4 & 62.1 &
        \textbf{70.7}\\
    \hline
    \end{tabular}
  }
  \caption{Experimental results on multi-hop open rule generation.}
    \label{table1}
\end{table}

\begin{table}[htp]
  \centering
   
   \resizebox{0.9\columnwidth}{!} %
  { 
  \renewcommand{\arraystretch}{1} %
  \setlength\tabcolsep{2.5pt} %
      \begin{tabular}{cccccc}
        \hline
        Scale & B1 & B2 & B4 & RL & Self-B2\\
        \hline
        774M+355M & 42.5 & 15.0 & 2.0 & 64.5 & 77.7\\
        774M+124M & 41.3 & 14.6 & 1.7 & 62.8 & 74.1\\
        355M+355M & 40.4 & 12.6 & 1.1 & 62.8 & 74.6\\
        355M+124M & 38.0 & 11.7 & 1.0 & 60.5 & 69.3\\
        \hline
    \end{tabular}
  }

   \caption{Comparative experiments on parameter size combinations for PRIMO subnetworks. The first scale number is the parameter size of \textit{Generation} module, and the second scale number is the parameter size of \textit{Extraction} module.}
    \label{table2}
\end{table}

\subsection{Main Results}
We report PRIMO's performance on the multi-hop open rule generation dataset in Table~\ref{table1}. We use BLEU-1/2/4 (B1, B2, B4)~\cite{DBLP:conf/acl/PapineniRWZ02} and ROUGE-L (RL)~\cite{Lin2004ROUGEAP} to evaluate whether he open rule chain generated by PRIMO is similar to the ground truth. We also report Self-BLEU-2 (Self-B2)~\cite{DBLP:conf/sigir/ZhuLZGZWY18}, which is used to measure diversity (smaller values indicate more diversity).

From the experimental results, it is evident that, apart from Vicuna-13B, PRIMO shows a significant performance improvement over other baseline models in the task of multi-hop open rule generation, especially when compared to Prompt and COMET.

\noindent \textbf{Comparison with PLM-based approaches} Firstly, in terms of rule quality and fluency, PRIMO achieves the best results in the BLEU-1, BLEU-4, and ROUGE-L metrics. Compared to the second-best model, Orion, PRIMO improves BLEU-1 by 4.4\%, BLEU-4 by 2.0\% and a significant 12\% improvement in ROUGE-L.

Secondly, in terms of rule diversity, we use the Self-BLEU-2 for evaluation. Looking at the results, PRIMO outperforms each of the baseline models on the Self-BLEU-2. Compared to Orion, PRIMO shows at least a 5.9\% improvement in the Self-BLEU-2. Hence, PRIMO generates significantly more diverse rules, suggesting that the use of a multi-stage network framework to extract implicit common sense knowledge from PLM is effective in enhancing the diversity of rules.

\noindent \textbf{Comparison with LLM-based approach} 
As shown in Table~\ref{table1}, it's evident that PRIMO achieves nearly the same experimental performance as Vicuna with a much smaller parameter size, indicating the effectiveness of our training strategy. The combination of domain fine-tuning and RLHF, together with the stacked architecture of small-scale models, can achieve the performance of LLM.%

\begin{figure}[t]
    \centering
    \includegraphics[width=0.9\columnwidth]{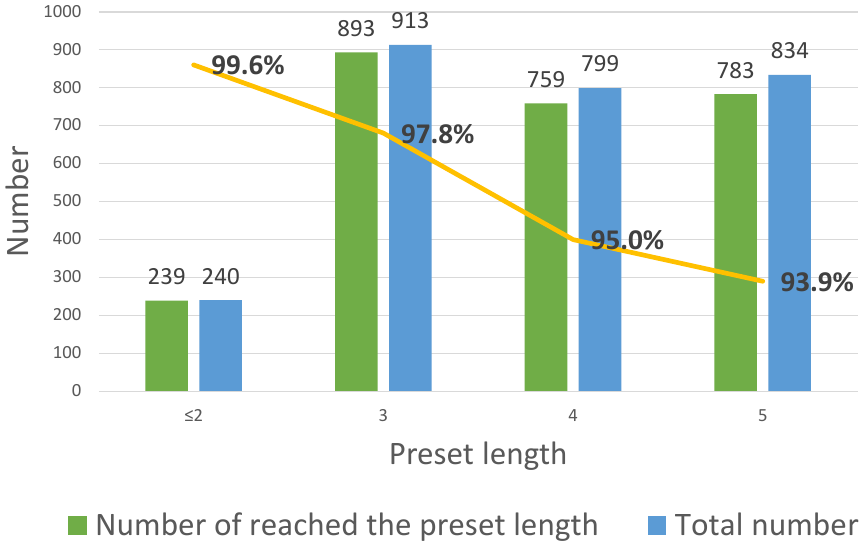}
    \caption{Statistics of length of rule chains. }
    \label{figure3}
\end{figure}

\begin{table}[t]
  \centering
  
   \resizebox{0.9\columnwidth}{!} %
  { 
  \renewcommand{\arraystretch}{1} %
  \setlength\tabcolsep{2pt} %
   \begin{tabular}{ccccc}
        \hline
        Threshold & Prompt & COMET & Orion & PRIMO\\
        \hline
        80\% & 37.8 & 56.7 & 43.1 & \textbf{21.6}\\
        90\% & 29.9 & 55.0 & 40.6 & \textbf{12.7}\\
        95\% & 28.3 & 54.5 & 40.1 & \textbf{10.9}\\
        \hline
    \end{tabular}
  }
  \caption{Comparison of rule atomic repetition rates.}
    \label{table3}
\end{table}

\noindent \textbf{Ablation Analysis} To determine how RLHF affects PRIMO performance, we conduct ablation experiments. We tested PRIMO without RLHF, denoted as PRIMO-without RLHF, which only underwent fine-tuning without RLHF. Additionally, we compared PRIMO with RLHF applied only to the \textit{Generation} module (PRIMO-train G\_Net) and only to the \textit{Extraction} module (PRIMO-train E\_Net) to explore the performance impact of RLHF on different stages of PRIMO. The results are shown in Table~\ref{table1}.

Firstly, we observe that applying reinforcement learning only to the \textit{Generation} module leads to an improvement in performance compared to the model without RLHF. This validates the effectiveness of the \textit{Generation} module in the whole rule generation process. It also confirms that using the \textit{Ranking} module as the reward model is effective for improving performance and underscores that the \textit{Ranking} module's scoring reflects human preferences. Additionally, applying reinforcement learning solely to the \textit{Extraction} module leads to a decrease in performance, while RLHF applied to both the \textit{Generation} and \textit{Extraction} module improves performance. It suggests a strong correlation between the \textit{Extraction} and \textit{Generation} modules, with the extraction results relying on the text generated by the \textit{Generation} module.

\begin{table*}[t]
    \centering
    \resizebox{\linewidth}{!}{
    \renewcommand{\arraystretch}{1.3}
    \begin{tabular}{>{\centering\arraybackslash}m{2.3cm}>{\centering\arraybackslash}m{7.8cm}>{\centering\arraybackslash}m{8.8cm}}
        \hline

        \centering \textbf{Premise atom}\strut
        & \centering \textless{}A\textgreater{} is stop of \textless{}B\textgreater{} \\{[}\textless{}A\textgreater{}:' Transit Stop ', \textless{}B\textgreater{}:' Transit Line'{]}\strut
        & \centering \textless{}A\textgreater{} is political party of \textless{}B\textgreater{} \\{[}\textless{}A:\textgreater{} 'Organization',\textless{}B\textgreater{}:'Form of Government'{]}\strut\tabularnewline
        \hline

        \multirow{4}{*}{\textbf{Orion}} & \centering \textless{}MASK\textgreater{} is a subway station on\textless{}MASK\textgreater{} & \textless{}MASK\textgreater{} is the legislative body of \textless{}MASK\textgreater{}\\
        & \centering \textless{}MASK\textgreater{} is a major part of\textless{}MASK\textgreater{} & \textless{}MASK\textgreater{} is the upper house of\textless{}MASK\textgreater{}\tabularnewline
        & \centering \textless{}MASK\textgreater{} is served by \textless{}MASK\textgreater{} & \textless{}MASK\textgreater{} is the upper house of \textless{}MASK\textgreater{}\tabularnewline
        & & \centering \textless{}MASK\textgreater{} is the upper house of \textless{}MASK\textgreater{}\tabularnewline
        \hline

        \multirow{4}{*}{\textbf{PRIMO}} & \centering \textless{}MASK\textgreater{} serve as a stop for \textless{}MASK\textgreater{}'s transit line & \textless{}MASK\textgreater{} directs \textless{}MASK\textgreater{}'s political development\tabularnewline
        & \centering \textless{}MASK\textgreater{} provides transit connections to \textless{}MASK\textgreater{} & \textless{}MASK\textgreater{} is the center of power of \textless{}MASK\textgreater{}\tabularnewline
        & \centering \textless{}MASK\textgreater{} serves as a primary stop for \textless{}MASK\textgreater{} & \textless{}MASK\textgreater{} may support \textless{}MASK\textgreater{}'s goals, objectives, and policies through its external influence operations\tabularnewline
        & & \centering \textless{}MASK\textgreater{} is the governing party of \textless{}MASK\textgreater{}\tabularnewline
        \hline
        
        \multirow{4}{*}{\textbf{Open rule chain}} & \centering \textless{}MASK\textgreater{} and \textless{}MASK\textgreater{}are interdependent & \textless{}MASK\textgreater{} is affiliated with or has a connection to \textless{}MASK\textgreater{}\tabularnewline
        & \centering \textless{} MASK \textgreater{} is a source point on \textless{} MASK \textgreater{} & \textless{}MASK\textgreater{} and \textless{}MASK\textgreater{} share common goals\tabularnewline
        & \centering \textless{} MASK \textgreater{} serves \textless{} MASK \textgreater{} as one of its stops & \textless{}MASK\textgreater{} and\textless{}MASK\textgreater{} have a symbiotic relationship\tabularnewline
        & & \centering\textless{}MASK\textgreater{} represents and advocates for the values and principles of \textless{}MASK\textgreater{}\tabularnewline
        \hline
    \end{tabular}
    }
    \caption{Case study. In Case 1, Orion generates rules with factual errors while PRIMO does not, which demonstrates the performance improvement brought by the introduction of ontology information in PRIMO. In Case 2, Orion generates rules with semantic repetition while PRIMO does not, which demonstrates the effectiveness of PRIMO in reducing the semantic repetition of rule atoms.}
    \label{table4}
\end{table*}

\noindent \textbf{Comparison of parameter scale impacts} We compare the performance of PRIMO with different parameter sizes, and the results are shown in Table~\ref{table2}. When keeping the \textit{Extraction} module fixed and changing the parameter size of the \textit{Generation} module, using a model with a larger parameter size leads to a better result, which indicates that as the parameter size of PLM increases, it possesses richer implicit knowledge and stronger text representation capabilities, which contribute to higher quality rule generation. When keeping the \textit{Generation} module fixed and changing the parameter size of the \textit{Extraction} module, similarly, models with larger parameter sizes also produce better rules, but the performance improvement is less pronounced compared to the former. This suggests that the \textit{Generation} module has a greater impact on the final results in the process of open rule generation than the \textit{Extraction} module, highlighting the critical role of the internal implicit knowledge provided by PLM in rule generation. Additionally, the experimental results indicate that the rule diversity continuously increases as the network parameter size decreases. We attribute this phenomenon to the fact that smaller model parameters result in lower overfitting levels, which can reduce the model's excessive adaptation to specific patterns and structures in the training data, thus generating more diverse rule atoms. Comparing all baseline models, even with the smallest parameter size in PRIMO, it achieves performance comparable to the state-of-the-art baseline model Orion based on PLMs, demonstrating the effectiveness and superiority of our approach.

\subsection{Semantics Repetition}
In order to investigate the effect of PRIMO on the generation of repeated rule atoms, we conduct an experiment on rule repetition rates. We compute the Jaccard similarity \cite{DBLP:journals/ipm/HamersHHJKRV89} between the strings of each hypothesis atom in the open rule chain. If the similarity exceeds a certain threshold, the two atoms are considered duplicates, and the count of duplicate atoms is increased by one. The final repetition rate is calculated by dividing the number of duplicate atoms by the total number of rule atoms  generated.

The experimental results are shown in Table~\ref{table3}. We find that PRIMO outperforms the baselines overall, leading baselines by at least 16.2\% for different threshold settings, which demonstrates that updating information about generated atoms in the prompt when generating the rule atom for the next hop can significantly reduce the semantic repetition of atoms. Moreover, when the similarity threshold is increased from 80\% to 90\%, there is a significant improvement in the performance of PRIMO (8.9\% improvement) compared to COMET and Orion, where there is no significant difference in performance (change of less than 2.5\%). The Prompt-based method shows a smaller improvement (an improvement of less than 8\%). We argue that PRIMO's atom repetition rate decreases dramatically as the similarity threshold is increased, suggesting that atoms generated by PRIMO are rarely nearly identical to previously generated atom representations, and that the semantic richness is significantly higher than the baselines.

\subsection{The Length of Rule Chain}
In our experiment, PRIMO has 65 data points resulting in no hypothesis atoms generated (rule chain length of 0). And there are 112 data points categorized as partial failure, meaning that PRIMO could generate hypothesis atoms but didn't  reach the preset rule chain length. Except for the chains with a length of 0, we analyze the length of the other generated rule chains, as shown in the Figure~\ref{figure3}. We observe that PRIMO performs well in generating shorter rule chains (hop \( \leq \) 3), but experiences a slight performance drop when trying to generate longer rule chains. We analyze that one reason is that it's genuinely impossible to infer more unknown relations between entity pairs. However, it also suggests that there is room for improvement in PRIMO's ability to generate longer rule chains.

\subsection{Case Study}
To visualize the generation quality of PRIMO, we provide some examples in Table~\ref{table4}.
When given the atom \texttt{<A> is stop of <B>}, although Orion generates some reasonable rule atoms, such as \texttt{<$\mathrm{MASK}$> is a major part of <$\mathrm{MASK}$>}, \texttt{<$\mathrm{MASK}$> is served by <$\mathrm{MASK}$>}, there are also some incorrect atoms. For instance, the first generated atom classified the entity as a subway station, but in reality, the correct answer should be a bus station. It is evident that Orion lacks modeling of entity ontology information related to rules, which leads to generation of wrong atoms. In comparison, PRIMO accurately describes the rule atom to the \texttt{Transit Stop} class, by introducing entity type information, such as \texttt{<$\mathrm{MASK}$> provides transit connections to <$\mathrm{MASK}$>}. Note that the rules generated by PRIMO excel in semantic expression, with each atom having corresponding synonymous expressions to the ground truth atoms, which indicates that PRIMO effectively improves the correctness of the final results by using the \textit{Generation} module to generate descriptive text related to the premise atoms.
When the given atom is \texttt{<A> is political party of <B>}, the generated atoms of Orion are restricted to those related to ``\texttt{is the legislature of}'', resulting in semantic repetition. On the other hand, PRIMO generates rule atoms that are both non-repetitive and reasonable. These cases show that PRIMO reduces the repetition rate of rule atoms by updating the information of previously generated atoms. Additionally, the preset rule chain length is 4, but Orion generates three identical atoms, indicating that Orion cannot generate rule atom after two hops. In contrast, PRIMO successfully reaches the preset length.

\section{Conclusion}
We propose a progressive multi-stage open rule generation method called PRIMO, i.e., generation-extraction-ranking. The latter module further refines and validates the results of the former module, which significantly enhances the semantic consistency of rule generation and mitigates duplicate generation. We optimize PRIMO with human feedback to further improve multi-hop open rule generation accuracy. Experiments show that PRIMO outperforms the PLM-based baseline model, and achieves nearly the same performance of LLMs using less than a tenth of parameters of a LLM.

\section{Acknowledgements}
This work is supported by the Natural Science Foundation of China (Grant No. U21A20488, 62206053). We thank the Big Data Computing Center of Southeast University for providing the facility support on the numerical calculations in this paper.


%
\end{document}